\documentclass[11pt]{article}
\usepackage[letterpaper,margin=1in]{geometry}
\usepackage[round,authoryear]{natbib}
\usepackage[utf8]{inputenc} 
\usepackage[T1]{fontenc}    
\usepackage{hyperref}       
\hypersetup{hidelinks}
\usepackage{url}            
\usepackage{booktabs}       
\usepackage{amsfonts}       
\usepackage{microtype}      
\usepackage{xcolor}         
\usepackage{graphicx}
\usepackage{subcaption} 
\usepackage{amsmath}
\usepackage{amssymb}
\usepackage{bm}
\title{Refined Grid Sampling for Efficient Implicit Neural Representations of Large-Scale Scientific Data}
\date{}

\author{%
  Guang Zhao\textsuperscript{1} \quad
  Xihaier Luo\textsuperscript{1} \quad
  Sean McCue\textsuperscript{3} \\
  Luke P. Van Roekel\textsuperscript{2} \quad
  Seungjun Lee\textsuperscript{1} \quad
  Yihui Ren\textsuperscript{1} \\
  Shinjae Yoo\textsuperscript{1} \quad
  Wei Xu\textsuperscript{1} \\
  {\rmfamily \textsuperscript{1}Brookhaven National Laboratory Upton, NY 11973, USA} \\
  {\rmfamily \textsuperscript{2}Los Alamos National Laboratory Los Alamos, NM 87545, USA} \\
  {\rmfamily \textsuperscript{3}University of California San Diego La Jolla, CA 92093, USA} \\
  {\small\texttt{gzhao@bnl.gov, xluo@bnl.gov, skmccue@ucsd.edu}} \\
  {\small\texttt{lvanroekel@lanl.gov}} \\
  {\small\texttt{slee19@bnl.gov, yren@bnl.gov, sjyoo@bnl.gov, xuw@bnl.gov}}
}

\begin{document}

\maketitle

\begin{abstract}
  Implicit neural representations (INRs) offer a powerful framework for modeling high-dimensional physical fields, but their training remains inefficient because evaluating dense coordinate grids is computationally expensive. Existing coordinate-pruning strategies rely on local error functions that can oversample correlated points and incur substantial overhead from full-grid error evaluation. We introduce Refined Grid Sampling (RGS), which combines stratified grid sampling, error-based refinement, and dynamic grid-resolution adjustment. This design reduces spatial redundancy without evaluating errors over the full grid. On high-resolution Navier--Stokes fields, RGS achieves higher reconstruction accuracy within a fixed wall-clock budget and reaches the accuracy of full-grid training up to $6\times$ faster than the evaluated baselines.
\end{abstract}

\section{Introduction}

Implicit neural representations (INRs) have emerged as a powerful tool for representing complex physical fields in a continuous and differentiable manner~\cite{gao2022nerf,xie2022neural}. However, INR training remains a major bottleneck, especially in large-scale scientific workloads. In computational fluid dynamics, for example, simulations can involve millions of mesh points to resolve complex flow phenomena, making na\"ive INR training prohibitively expensive. This challenge motivates adaptive sampling methods that retain INR fidelity while reducing computational cost, thereby enabling broader use of INRs in scientific discovery~\cite{cole2023spatial, luo2024continuous, molaei2023implicit, zhong2021cryodrgn2}.

Several pruning and sampling methods address this challenge by using only a subset of coordinates in each training iteration, thereby reducing the number of points used in backpropagation. Examples include Implicit Neural Teaching (INT)~\cite{zhang2024nonparametric}, Soft Mining~\cite{kheradmand2024accelerating}, and EVOlutionary Selector (EVOS)~\cite{zhang2025evos}. These approaches typically use local scores based on prediction errors, under the assumption that coordinates with larger errors should receive priority during training. Although effective in principle, such methods face two key limitations. First, \textit{they ignore spatial redundancy}: physical continuity means that adjacent points often carry redundant information. Selecting only the points with the highest errors can oversample dense clusters instead of producing a diverse and representative subset. Second, \textit{they introduce significant computational overhead}. Identifying high-error points can require a full forward pass over all coordinates. This cost can match or exceed the savings from backpropagating through only a small subset, negating the computational gains.
To address these limitations, we introduce \textit{Refined Grid Sampling (RGS)}, a two-stage strategy that combines grid sampling with error-based selection. RGS first promotes spatially representative coverage by sampling candidate coordinates from grid cells, then refines the candidate set by prioritizing high-error points. This design reduces redundancy and avoids exhaustive error computation over the full domain. To accommodate decreasing spatial correlations during training, we progressively refine the grid resolution. Experiments on Navier--Stokes vorticity fields demonstrate that RGS accelerates INR training while improving reconstruction accuracy over the evaluated baselines.

\section{Method}
\subsection{Problem Definition}
We formalize coordinate pruning for implicit neural representation (INR) training. An INR is a neural network parameterized by $\theta$ that maps spatial coordinates to continuous field values, $f_\theta: \mathcal{X} \rightarrow \mathcal{Y}$. Here, $\mathcal{X} \subset \mathbb{R}^d$ denotes the input domain and $\mathcal{Y}\subseteq \mathbb{R}^c$ denotes the output space. The objective is to approximate a target signal $u: \mathcal{X} \rightarrow \mathcal{Y}$. Conventional training minimizes the full loss
\begin{equation}
\mathcal{L}(\theta) = \frac{1}{N} \sum_{i=1}^N \| f_{\theta}(\bm{x}_i) - u(\bm{x}_i) \|_2^2,
\end{equation}
which requires evaluating all coordinates and is computationally prohibitive when $N$ is large. Coordinate pruning instead restricts optimization to a subset of coordinates at each iteration. At training step $t$, a subset $\mathcal{S}_t \subset \mathcal{X}$ of size $|\mathcal{S}_t| \ll N$ is selected, and optimization proceeds with the pruned loss
\begin{equation}
\mathcal{L}'(\theta) = \frac{1}{|\mathcal{S}_t|} \sum_{\bm{x}_i\in \mathcal{S}_t} \| f_{\theta}(\bm{x}_i) - u(\bm{x}_i) \|_2^2. 
\end{equation}
The challenge is to choose $\mathcal{S}_t$ so that training remains efficient while the final error $\mathcal{L}(\theta)$ stays low under a fixed computational budget. RGS constructs $\mathcal{S}_t$ using stratified coverage, error-based refinement, and dynamic grid adjustment.

\subsection{Stratified Grid and Error-Based Sampling}

RGS uses a two-stage sampling strategy that combines stratified grid coverage with error-based refinement. In the first stage, the input domain $\mathcal{X}$ is partitioned into equal-sized, non-overlapping grid cells, and a small number of candidate coordinates are drawn uniformly from each cell. This promotes coverage of the full field, avoids concentrating candidates in only a few regions, and limits the number of forward evaluations in the second stage.

In the second stage, error-based refinement further reduces the candidates used for backpropagation. For each sampled candidate, we compute the local prediction error $\|f_{\theta}(\bm{x}_i)-u(\bm{x}_i)\|_2$ and retain the global top-$k$ candidates with the largest errors. This step emphasizes informative points while selecting from a spatially stratified candidate pool.

Together, stratification and error refinement yield a subset $\mathcal{S}_t$ that balances diversity and informativeness, reducing both redundancy and computational overhead compared to purely error-driven selection.

\vspace{-2mm}
\subsection{Spatial Correlation Analysis and Dynamic Grid Size Adjustment}
\label{sec:correlation}

\begin{figure}[htbp]
\vspace{-2mm}
    \centering
    \begin{subfigure}[b]{0.62\textwidth}
        
        \centering
        \includegraphics[width=\textwidth]{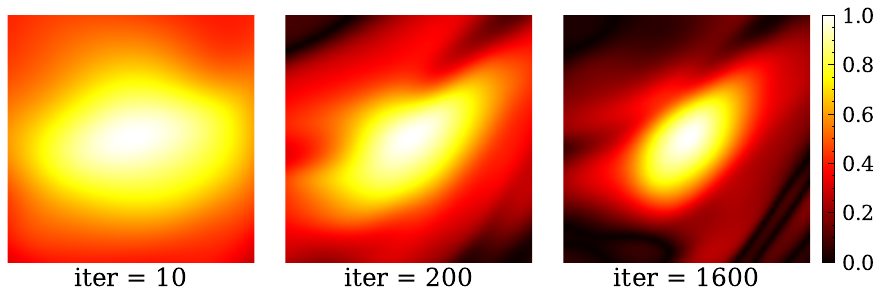}
        \label{fig:heatmaps}
    \end{subfigure}
    \hfill
    \raisebox{2mm}{
    \begin{subfigure}[b]{0.34\textwidth}
        
        \centering
        \includegraphics[width=\textwidth, height=3.1cm, keepaspectratio]{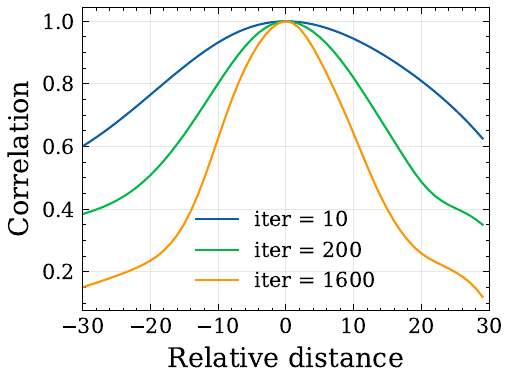}
        \label{fig:cross_section}
    \end{subfigure}}
    \caption{Evolution of gradient cosine correlation during INR training.
\textit{Left:} Heatmaps of the correlation between a reference coordinate and its neighbors within a $60\times60$ region at training epochs $t=10$, 200, and 1,600. The high-correlation area contracts during training, indicating reduced spatial redundancy. \textit{Right:} Center-row correlation profiles, showing sharper decay at later epochs.}
    \vspace{-2mm}
    \label{fig:correlation_analysis}
\end{figure}

The core idea of stratified grid sampling is to represent each cell's contribution to parameter updates using only a few coordinates, thereby reducing redundancy while preserving fidelity. This assumption is most appropriate when the gradients of per-coordinate losses within a cell are highly correlated, meaning that neighboring points provide similar training signals for the parameters $\theta$.

To examine how gradient correlation evolves during training, we analyze an INR trained on a $1024\times1024$ Navier--Stokes field using all coordinates. At epochs 10, 200, and 1,600, we compute the per-coordinate loss gradient, $\nabla_{\theta}\ell(\bm{x};\theta)=\nabla_{\theta}\|f_{\theta}(\bm{x})-u(\bm{x})\|_2^2$, and measure its cosine correlation between a central coordinate and neighboring coordinates within a $60\times60$ region. Figure~\ref{fig:correlation_analysis} shows how this correlation decays over space and training time.

From the figure, during the early stages of training, we observe that the per-pixel loss gradients exhibit high spatial correlation over large regions, whereas in later stages this correlation becomes increasingly localized. This behavior can be interpreted through the \textit{lens of spectral bias} (or \textit{frequency principle}) of neural networks, which states that, under gradient-based optimization, neural networks tend to learn low-frequency (smooth) components of the target function before fitting high-frequency details \cite{rahaman2019spectral, xu2019frequency}.

\begin{figure}[htbp]
    \centering
    \includegraphics[width=0.95\linewidth]{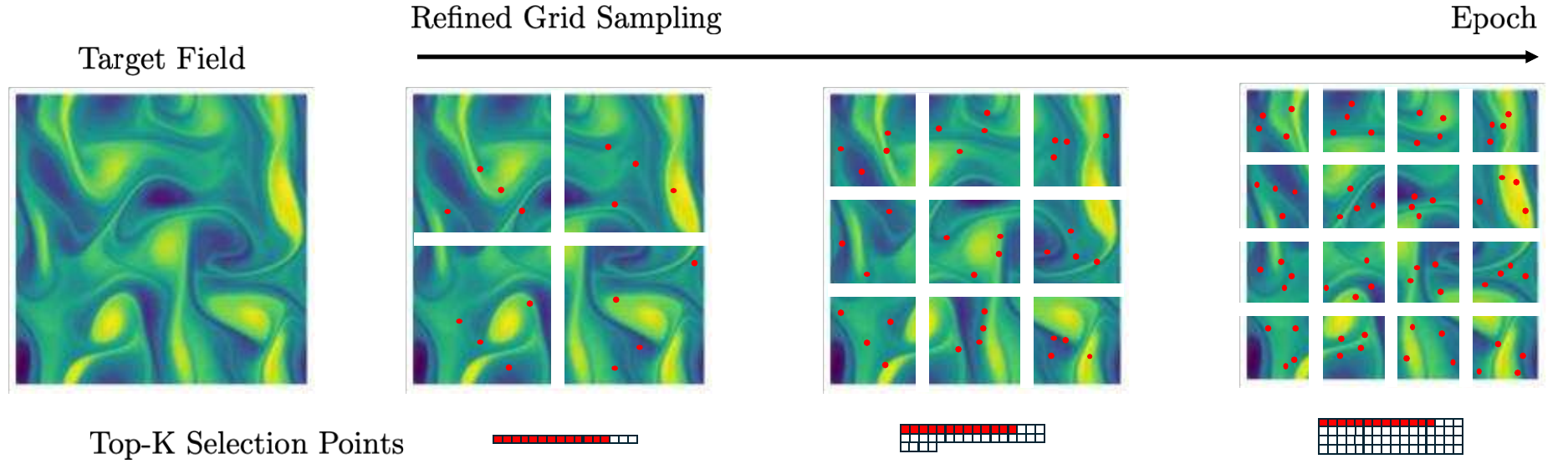}
    \caption{Refined Grid Sampling (RGS) framework. RGS combines stratified grid sampling with error-ranked selection to balance spatial coverage and informative points, reducing redundancy and forward-pass cost. Sampling starts with coarse grids and progressively refines as gradient correlations decay, improving reconstruction accuracy.}
    \label{fig:diagram}
    \vspace{-3mm}
\end{figure}

Based on these observations, we adjust the grid resolution during training. Coarser grids are used when gradients are highly correlated in the early stages, and finer grids are introduced as correlations decay and gradients become more localized. In practice, we specify the starting and ending numbers of grid cells per spatial axis and interpolate them linearly over training epochs. As the grid is refined, the first stage yields more candidates for error evaluation, while the second stage keeps the selected top-$k$ size fixed for backpropagation. Although finer grids increase the number of evaluated coordinates, that number remains far below the full-grid size. Appendix~\ref{sec:appendix_efficiency} shows that RGS runtime per epoch remains nearly constant during training. Figure~\ref{fig:diagram} provides an overview of RGS.

\section{Experiments}
\label{sec:experiments}

\paragraph{Dataset.}
We consider the 2D incompressible Navier--Stokes equation in vorticity form on the unit torus with periodic boundary conditions:
\begin{equation}
    \partial_t \omega(\bm{x},t) + u(\bm{x},t) \cdot \nabla \omega(\bm{x},t) = \nu \Delta \omega(\bm{x},t) + f(\bm{x}), \quad
    \nabla \cdot u(\bm{x},t) = 0, \quad \omega(\bm{x},0) = \omega_0 (\bm{x}),
\end{equation}
where $u$ is the velocity field, $\omega=\nabla\times u$ is the vorticity field, and $\omega_0$ is the initial condition. We use Reynolds number $\mathrm{Re}=1/\nu=1000$ and a $1024\times1024$ spatial grid. The external forcing is $f(\bm{x})=-4\cos(4x_2)-0.1\omega(\bm{x},t)$, where $x_2$ is the vertical coordinate; the drag term prevents energy accumulation at large scales~\cite{drag_force,lee2024inducing}. This setup produces chaotic, multiscale turbulence and provides a challenging reconstruction benchmark.

\paragraph{Baselines and evaluation protocol.}
We use a SIREN model~\cite{sitzmann2020implicit} with approximately 300,000 parameters. We compare: (1) \textbf{Full Training}, which uses every coordinate; (2) \textbf{Random}, which samples coordinates uniformly; (3) \textbf{INT}~\cite{zhang2024nonparametric}, which selects coordinates according to prediction error; (4) \textbf{EVOS}~\cite{zhang2025evos}, an evolutionary selector guided by prediction error; and (5) \textbf{RGS}. For every sampling method, the subset used for backpropagation contains 0.1\% of the full coordinate grid. Full Training uses a learning rate of $5\times10^{-4}$, while the sampling methods use $1\times10^{-4}$, as selected by hyperparameter tuning. We independently reconstruct six time frames from the Navier--Stokes trajectory and report the mean and standard deviation across these runs. Reconstruction metrics are evaluated on the complete spatial grid.

\begin{figure}[htbp]
    \centering
    
    \begin{subfigure}[t]{0.47\textwidth}
        \centering
        \includegraphics[width=\textwidth]{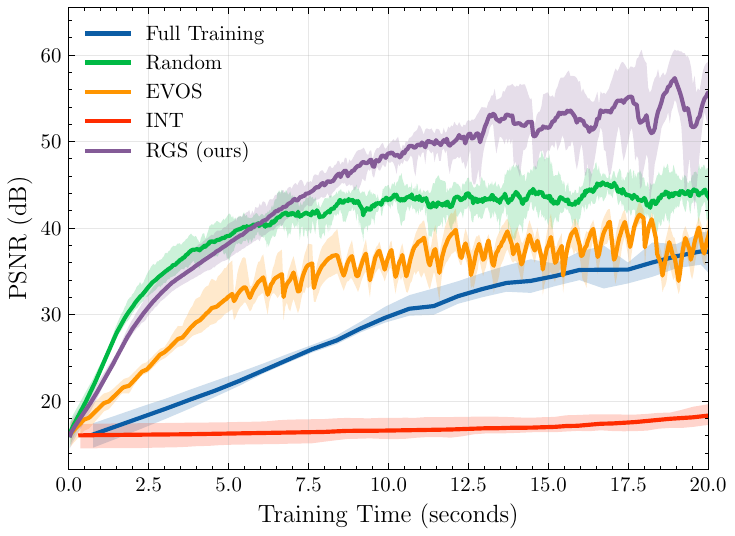}
        \caption{PSNR as a function of wall-clock training time. RGS converges faster than Random, EVOS, and INT and surpasses the accuracy reached by Full Training within the evaluated budget.}
    \end{subfigure}
    \hfill
    \begin{subfigure}[t]{0.50\textwidth}
        \centering
        \includegraphics[width=\textwidth]{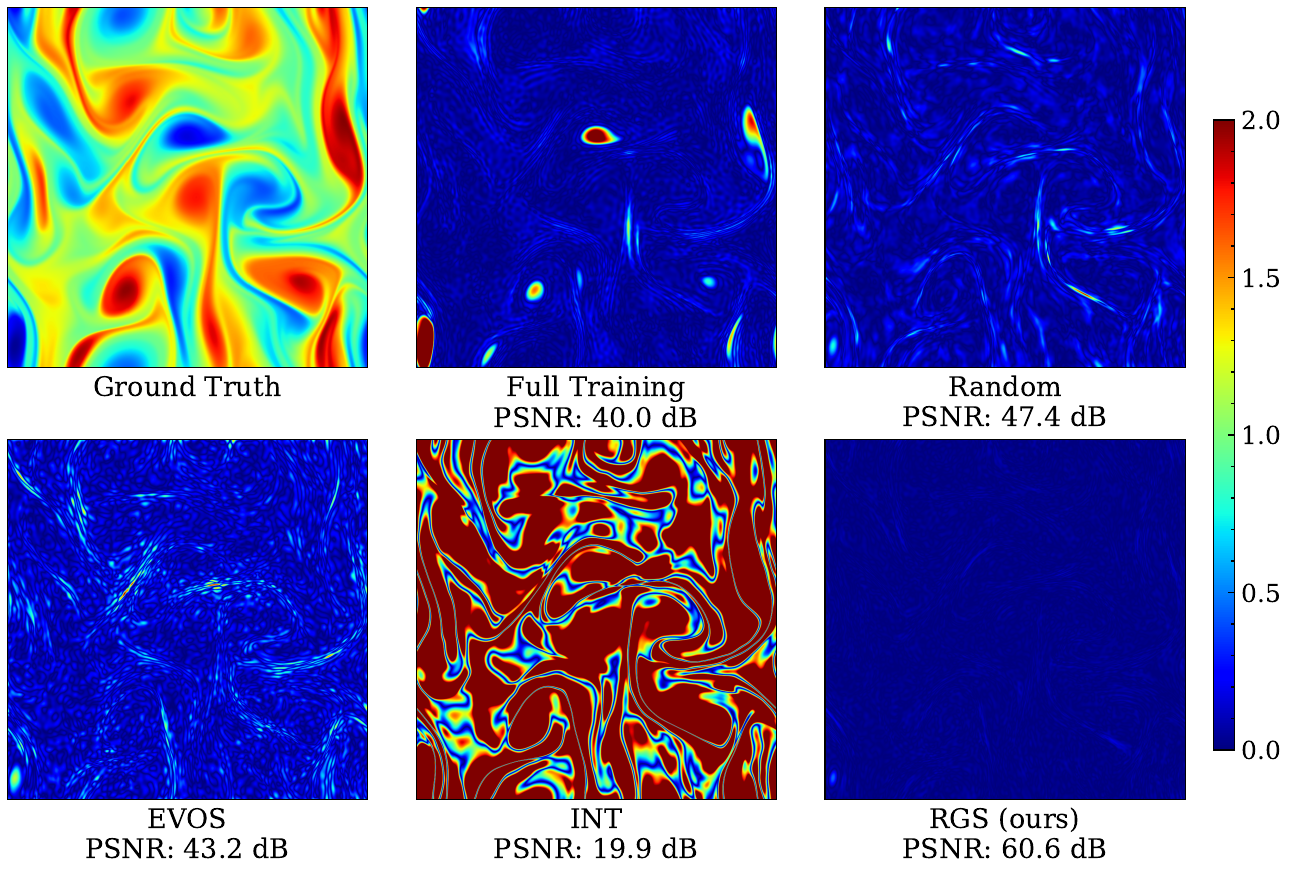}
        \caption{Error maps after a 20-second wall-clock budget. The top-left panel shows the ground truth; the remaining panels show absolute reconstruction error for each method. The $[0,2.0]$ color scale applies only to the error maps.}
    \end{subfigure}

    \caption{PSNR and error-map comparison of sampling strategies on Navier--Stokes fields.}
    \label{fig:ab}
\end{figure}

\begin{table}[ht]
\centering
\small
\begin{tabular}{@{}lcccc@{}}
\toprule
\textbf{Sampling strategy} & \textbf{PSNR} & \textbf{SSIM} & \textbf{Rel. RMSE} & \textbf{Time/epoch (s)} \\
\midrule
Full Training & 37.19 $\pm$ 2.29 & 0.978 $\pm$ 0.024 & 0.092 $\pm$ 0.031 & 0.190 \\
Random & 43.72 $\pm$ 1.20 & 0.982 $\pm$ 0.004 & 0.042 $\pm$ 0.006 & \textbf{0.003} \\
EVOS & 37.41 $\pm$ 1.70 & 0.930 $\pm$ 0.017 & 0.087 $\pm$ 0.017 & 0.006 \\
INT & 18.38 $\pm$ 0.85 & 0.212 $\pm$ 0.094 & 0.773 $\pm$ 0.109 & 0.036 \\
RGS (ours) & \textbf{56.24 $\pm$ 2.73} & \textbf{0.998 $\pm$ 0.001} & \textbf{0.010 $\pm$ 0.003} & 0.005 \\
\bottomrule
\end{tabular}
\caption{Performance after a 20-second wall-clock budget, reported as mean $\pm$ standard deviation over six independently reconstructed time frames. Best values are shown in bold.}
\label{tab:method_comparison}
\end{table}

Figure~\ref{fig:ab} and Table~\ref{tab:method_comparison} summarize the reconstruction results. RGS reaches the accuracy attained by Full Training up to $6\times$ sooner. Under the common 20-second budget, RGS obtains the best reconstruction accuracy across all metrics, reaching 56.24 dB PSNR and 0.998 SSIM. Its error map also contains visibly smaller errors. RGS remains close to the lightest baselines in per-epoch cost: its 0.005-second epoch time is slightly above Random (0.003 seconds) and substantially below INT (0.036 seconds). These results indicate that stratified, correlation-aware sampling improves fidelity while retaining low training cost.
\section{Conclusion}

We introduced Refined Grid Sampling (RGS) for efficient INR training. RGS combines stratified coverage, error-based refinement, and dynamic grid adjustment to address the spatial redundancy and computational overhead of existing coordinate-pruning methods. Experiments on high-resolution Navier--Stokes fields show that RGS improves reconstruction quality under a fixed wall-clock budget and reaches full-training accuracy up to $6\times$ faster. These results demonstrate the potential of structured coordinate sampling for scalable and efficient modeling of complex physical systems.

\section*{Acknowledgments}
\textbf{Funding.} This material is based upon work supported by the U.S. Department of Energy, Office of Science, Office of Advanced Scientific Computing Research and Office of Biological and Environmental Research, via the RAPIDS Institute under the Scientific Discovery through Advanced Computing (SciDAC) program and the project of Improving Projections of AMOC and its Collapse Through Advanced Simulations (ImPACTS), under Contract No. DE-SC0012704 and FWP No. CC122 and CC127, and award B\&R-KJ0401010 and FWP-CC147.\\

\bibliographystyle{plainnat}
\bibliography{inr_sampling}


\appendix

\section{Additional Efficiency Results}
\label{sec:appendix_efficiency}

\begin{figure}[htbp]
    \centering
    \includegraphics[width=0.7\textwidth]{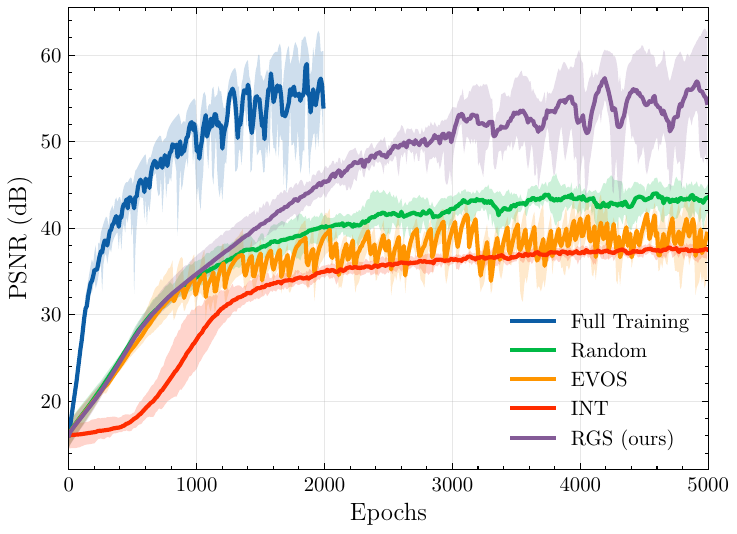}
    \caption{PSNR versus training epochs for different sampling strategies. RGS converges faster and attains higher final accuracy than the error-driven baselines, while random sampling converges more slowly.}
    \label{fig:epoch-psnr}
\end{figure}

\begin{figure}[htbp]
    \centering
    \includegraphics[width=0.7\linewidth]{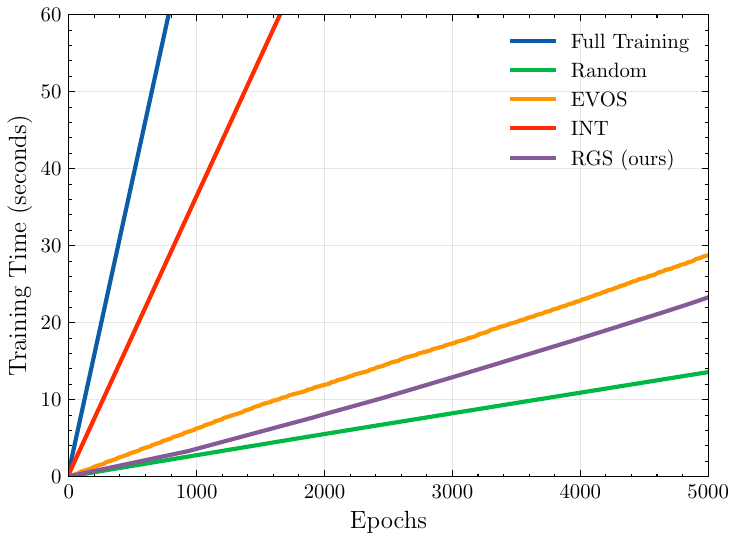}
    \caption{Training time versus epochs for different sampling strategies. RGS substantially reduces training time relative to full training and INT, while remaining competitive with random sampling and EVOS.}
    \label{fig:time-epoch}
\end{figure}

For a more detailed analysis of computational efficiency, Figures~\ref{fig:epoch-psnr} and~\ref{fig:time-epoch} compare convergence and cumulative runtime over epochs. These plots complement the main results by showing convergence over epochs and how runtime accumulates during training. As discussed in Section~\ref{sec:experiments}, RGS achieves a favorable trade-off between convergence speed and computational cost.

\end{document}